\documentclass[conference]{IEEEtran}
\IEEEoverridecommandlockouts
\usepackage{cite}
\usepackage{amsmath,amssymb,amsfonts}
\usepackage{algorithmic}
\usepackage{algorithm}
\usepackage{graphicx}
\usepackage{textcomp}
\usepackage{xcolor}
\usepackage{balance}
\def\BibTeX{{\rm B\kern-.05em{\sc i\kern-.025em b}\kern-.08em
    T\kern-.1667em\lower.7ex\hbox{E}\kern-.125emX}}
    
\begin{document}

\title{DAGCN: Dual Attention Graph Convolutional Networks}

\author{\IEEEauthorblockN{Fengwen Chen \IEEEauthorrefmark{1}, Shirui Pan\IEEEauthorrefmark{2}, Jing Jiang\IEEEauthorrefmark{1}, Huan Huo\IEEEauthorrefmark{3} Guodong Long\IEEEauthorrefmark{1}}
\IEEEauthorblockA{\IEEEauthorrefmark{1} Centre for Artificial Intelligence, FEIT, University of Technology Sydney, Australia \\ \IEEEauthorrefmark{2} Faculty of Information Technology Monash University, Australia\\ \IEEEauthorrefmark{3} School of software, FEIT, University of Technology Sydney, Australia\\
Email: fengwen.chen@student.uts.edu.au, shirui.pan@monash.edu, jing.jiang@uts.edu.au \\ huan.huo@uts.edu.au, guodong.long@uts.edu.au}
}

\maketitle

	\begin{abstract}
        Graph convolutional networks (GCNs) have recently become one of the most powerful tools for graph analytics tasks in numerous applications, ranging from social networks and natural language processing to bioinformatics and chemoinformatics, thanks to their ability to capture the complex relationships between concepts. At present, the vast majority of GCNs use a neighborhood aggregation framework to learn a continuous and compact vector, then performing a pooling operation to generalize graph embedding for the classification task. These approaches have two disadvantages in the graph classification task: (1)when only the largest sub-graph structure ($k$-hop neighbor) is used for neighborhood aggregation, a large amount of early-stage information is lost during the graph convolution step; (2) simple average/sum pooling or max pooling utilized, which loses the characteristics of each node and the topology between nodes. In this paper, we propose a novel framework called, dual attention graph convolutional networks (DAGCN) to address these problems. DAGCN automatically learns the importance of neighbors at different hops using a novel attention graph convolution layer, and then employs a second attention component, a  self-attention pooling layer, to generalize the graph representation from the various aspects of a matrix graph embedding. The dual attention network is trained in an end-to-end manner for the graph classification task. We compare our model with state-of-the-art graph kernels and other deep learning methods. The experimental results show that our framework not only outperforms other baselines but also achieves a better rate of convergence.
	\end{abstract}

	\section{Introduction}

	     \par Graph structured or network data are rapidly becoming ubiquitous in our daily lives, e.g., World Wide Web network, transportation networks, and protein interaction networks. Researchers have conducted extensive research on many important machine learning applications in graph with both supervised and unsupervised fashion \cite{bacciu2018contextual}, such as vertex classification\cite{chuang1990three}, anomaly detection\cite{noble2003graph}, link prediction\cite{li2018diffusion} and recommendation system\cite{fouss2007random}, but the complexity of graph data imposes great challenges for many tasks including one of the central tasks in the field, graph classification (but not $node$ classification), which aims to assign a class label to an entire graph. In a cheminformatics dataset, for instance, atoms are represented by graph nodes and chemical bonds are represented by graph edges. A graph classification model can be applied to a dataset for many applications, from detecting molecular status, such as cancer activity detection or solubility detection, molecular properties, such as toxicity detection.
	   
	     To solve the problem of graph classification, the most widely used strategy consists of graph statistic-based methods which are able to represent the graph in various aspects. Graph kernel\cite{vishwanathan2010graph,shervashidze2011weisfeiler} is the most popular of these techniques; it employs a kernel function to measure the positive semi-definite graph similarity between pairs of graphs \cite{shervashidze2009efficient}. The classification task can then be conducted on a similarity matrix by using supervised algorithms like Support Vector Machine \cite{cortes1995support}. By decomposing the graph into sub-structures, the graph kernel is capable of directly processing the graph data without transforming it into feature vectors. As a result, it has achieved dramatic success in node classification, link prediction, node clustering and so on.
	   
	     Graph kernel-based algorithms nevertheless still suffer from natural limitations, such as the exponential growth of computation operations and the fixed feature design, which will be discussed in more detail in Section IV. 
	     Other algorithms \cite{pan2016joint} attempt to distinguish and select the sub-graph features for graph classification by recursively applying an aggregation process on each node with the attributes from local neighbors to learn the node representations. The graph feature is then generated according to all the learned node representations in the graph.

	     Deep learning-based approaches like graph neural network have also been applied diffusely for network representation. These approaches embed the given graph and the side information associated with it into a continuous and compact vector space. After embedding, the graphs sharing common patterns are expected to be close to each other in the vector space, therefore classical machine learning methods can be applied to the embedded vector for graph classification. However, while the graph-structured data preserves more relational information than other data formats, it also incurs more complicated noise. How to learn a good representation while screening out the interference caused by the complex noise of each node in a graph has become a significant challenge. Moreover, sub-graphs which consist of multiple nodes, or even the entire graph are required in the graph classification task to achieve a more comprehensive analysis. Hence, obtaining the graph representation based on node representation is another non-negligible challenge.
	     
	     \begin{figure*}
 	        \centering
 	        \includegraphics[width=1\textwidth]{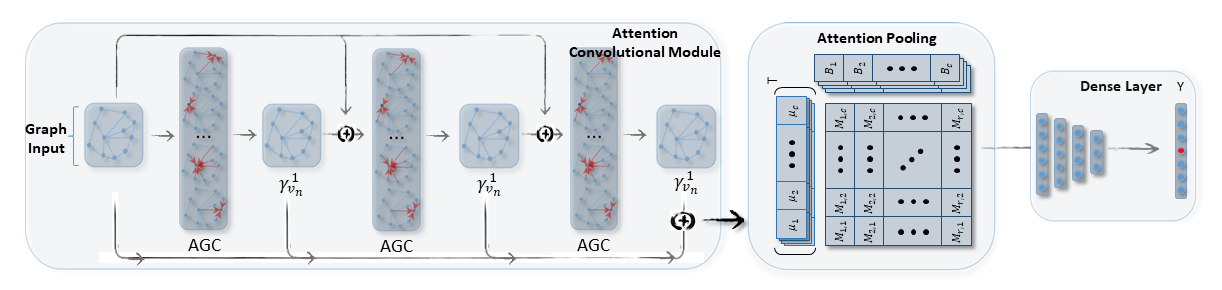}
 	        \caption{The architecture of the dual attention graph convolution network (DAGCN). The model consists of three parts: (1) The left tier is the attention graph convolution module with three AGC layers ($m=3$) which learns the hierarchical local substructure features by aggregating the hops of its neighbors. (2) The middle part is the attention pooling layer, the matrix B is the attention coefficient matrix. (3) The final graph embedding matrix $M$ is then sent to a dense layer for final predictions.}   
 	        \label{fig:architecture}
 	        
	    \end{figure*}
	     
	     Many researches have concentrated on re-factorizing neural network architectures to directly process structured graph data\cite{monti2017geometricb,pan2016tri,defferrard2016convolutional,velickovic2017graph,pan2015finding}. However, graph data are complex in many ways; for example, the topological structure information of different sub-graphs is fickle when the size is varied. Most existing graph neural network frameworks are limited by two factors when dealing with this scattered information because: 1) these frameworks ignore the significance of different hop neighbors. Only the final aggregation output is used, i.e., only the largest sub-graph is used to learn the node representation. 2) they mainly apply average/sum pooling or max pooling which fails to leverage the valuable information of a node or sub-graph in the graph. While conducting graph classification, we attempt to pay more attention to the graph signature \cite{lee2018graph} (i.e. the special node or sub-graph), which is only a small segment of the entire graph. In contrast, a simple average/sum pooling or max pooling could result in a model that is constructed on too much irrelevant information. 
	     
	     To address the above problems, we propose a novel framework named Dual Attention Graph Convolution Network (DGCNN). The core idea of the proposed DGCNN is to identify and maximize the importance of the nodes or sub-graph when conducting graph classification. We first merge the attention technique in the graph convolution operation to capture the arbitrary local structure information in a graph. A self-attention pooling layer then generates an adaptive combination representation matrix, in which each row in the learned matrix represents one perspective of the graph. Our contributions in this paper are threefold: 
	     \begin{itemize}
	         \item We propose a novel attention graph convolution technique which is capable of leveraging the information from different hop neighbors rather than the k-hop only; 
	         \item We propose a novel graph self-attention pooling technique which extracts a more informative embedding matrix containing multiple significant nodes or sub-graphs; 
	         \item We conduct experiments and compare our method with both deep learning-based methods and graph kernel-based algorithms. The experiment results demonstrate that the proposed Dual Attention Graph Convolutional Network (DAGCN) outperforms the deep learning benchmarks for graph classification and are highly comparable with state-of-the-art graph kernels.
	     \end{itemize}
	
	\section{Related Work}
	    There have been many attempts on graph classification tasks in the literature. The earliest experiments can be traced back to 1998 when Frasconi et al. \cite{frasconi1998general} used a recursive neural network to process directed acyclic graphs. Subsequently, Gori et al.\cite{gori2005new} introduced Graph Neural Networks (GNNs) to extend the neural network for graph-structured data. GNNs normally consist of an aggregation process which aggregates the node features a certain number of times or until equilibrium is reached to produce an embedding for each node. This idea has been broadly adopted and improved in many tasks \cite{henaff2015deep,li2015gated,pan2018adversarially}.
	    
	
        With the great success of computer vision, there is an increasing interest in generalizing convolutions to the graph domain. Bruna et al.\cite{bruna2013spectral} first generalized the convolution operation to the graph's spatial domain after the original data have been transformed by Graph Fourier Transform (GFT). Since the computation of eigenvectors is involved, computational complexity has become a serious issue. Many researchers have worked on optimizing the convolution filters to reduce the computational complexity \cite{henaff2015deep,defferrard2016convolutional,li2018adaptive,levie2017cayleynets}. However, the learning process in all the aforementioned spectral approaches usually depends on the Laplacian eigenbasis, which handles the entire graph at one time. Thus, the issue of scalability and computational complexity still cannot be overcome.
        
        Duvenaud et al.\cite{duvenaud2015convolutional} introduced a spatial GCN that directly defines the convolutions on a graph without a transform. Each node propagates the features from its 1-hop neighbors to generate a differentiable fingerprint which simulates the circular fingerprints.After Kipf et al.\cite{kipf2016semi} simplified the concept, Atwood et al.\cite{atwood2016diffusion} extended this idea by propagating $n$ different hops to the center node with different weights. A common challenge of these approaches is how to define the range of neighborhoods to aggregate and the strategies for obtaining information from neighbors. More recently, Niepert et al. \cite{niepert2016learning} and Hamilton et al. \cite{hamilton2017inductive} addressed the challenge in another way by sampling a fixed-size neighborhood for each node and then performing the aggregation. Lately, Tran et al. \cite{van2018filter} further optimizing GCNs by extending the basic graph convolution operator. These approaches have achieved high levels of performance and have increased the scope of GCN applications. Given rapid developments in the field of GCN, we point readers to our recent, comprehensive review in  \cite{wu2019comprehensive}.
        
        An important component that usually comes with CNNs, the pooling layer, can also be generalized to graph-structured data. It is a down-sampling strategy that largely reduces the spatial size of the input while roughly retaining its location relationships. Mean pooling is the most commonly used graph pooling strategy due to its conciseness. Easily mean all node's information could also solve the issues of rotational invariance and yield better performance \cite{henaff2015deep}. To better preserve the relationship between nodes, Defferrard et al. \cite{defferrard2016convolutional} and Zhang et al. \cite{zhang2018end} proposed approaches that perform pooling after the nodes have been rearranged in a meaningful order using a different strategy. This could be viewed as selecting similar parts of different graphs so that the preserved node relation can be used effectively. Overall, the essence of pooling is to reduce the size of the input (usually the node representation) by losing some information. Deciding which information to retain is the key to the model. 
        
        Attention mechanisms have already become the standard in many fields for a number of tasks\cite{bahdanau2014neural,shen2017disan}. The most important advantage of attention mechanisms is that they are able to handle the variably sized inputs by focusing on the most relevant parts of the inputs to make decisions. When attention is implemented on the same input, it is called Self-Attention \cite{lin2017structured}. There is little literature on the topic of attention mechanisms on graph-structured data. Velickovic et al.\cite{velickovic2017graph} employed attention to dynamically compute the weight of each node's neighbors during aggregation. Attention mechanisms have mainly been used in aggregation processing. A few attempts have been made to extend attention beyond aggregation \cite{lee2018graph,abu2018watch}, but some issues have still never been studied. Inspired by recent works and the defect of them, we propose our model which uses an attention technique to maximize the use of information that underlies the original graph input.
	
	\section{Problem Definition and Framework}
	   A graph is represented as $g=(V_g, E_g, A_g, X_g)$, where $V_g$ is a vertex set {$v_i$}, $i$ = 1, ...., n. $E_g$ represents the linkages between nodes, denoted as $e_{i,j}=<v_i, v_j> \in E, i\neq j$. An unweighted adjacency matrix $A_g \in \lbrace 0, 1\rbrace^{N_i \times N_i}$ represents the graph's topological structure by setting $A_{i,j}=1$ if $e_{i,j} \in E_g$, otherwise $A_{i,j} = 0$. $N_i$ is the size of the graph $g_i$. $X \in \mathbb{R}^{n \times c}$ indicates the $c$ channel content features associated with each node $v_i$.
	    
	   Given a set of graphs $G = (g_1, g_2, \cdots g_n)$ with their labels $Y=(y_1, y_2, \cdots, y_n)$, the \textbf{goal} of our paper is to learn a function $f(g_i) \to y_i \in L$, where $L=\{c_1,\cdots, c_{|L|}\}$ is the class labels for the graphs. In this paper, we will develop a novel graph convolutional network which employs dual attentions at both node level and graph level, for graph classification.
	   
	   \subsection{Overall Framework}
	   Our objective is to learn a classifier which could classify the given graph $G$. To achieve this, we propose a novel dual attention graph convolution network (DAGCN). Figure \ref{fig:architecture} demonstrates the work-flow of DAGCN which consists of two modules: the attention graph convolution module and the attention pooling module.
	   
	   \begin{itemize}
            \item \textbf{Attention Graph Convolution Module} The attention graph convolution module is constructed of several attention graph convolution layers. Each layer takes the features $X$ and adjacency matrix $A$ to extract the hierarchical local substructure features of the vertices from different hops of neighbor. 
	        \item \textbf{Attention Pooling Layer} The attention pooling layer uses the nodes' embedding to learn multiple graph representation from different aspect and outputs a fixed size, matrix graph embedding.
	   \end{itemize}
	
	\section{Dual Attention Graph Convolution}
		The DAGCN consists of three parts: (1) the attention graph convolution module; (2) the self-attention pooling layer; and (3) the fully connected classifier. In this section, we first address the problem of traditional GCNs and, then propose our attention graph convolution module and self-attention pooling layer.
		
		\subsection{Traditional Graph Convolution} We start by describing the traditional graph convolution layer and then propose DAGCN to address the shortcomings. The most general form of graph convolution with depth of $k$ can be expressed recursively by a broadly followed convolution structure denoted as:

		\begin{eqnarray}
		H^{k+1} = \phi(\widetilde{A}\widetilde{D}^{-1}H^kW) \qquad 	 H^0=X,
		\label{eq:eq1}
		\end{eqnarray}
		
		where $\widetilde{A} = A + I_n$ is the adjacency matrix with self-connection for each node, $\widetilde{D}$ is the diagonal node degree matrix of $\widetilde{A}$, $\widetilde{A}\widetilde{D}^{-1}$ represents the normalized graph structure, and $W$ is the model parameter that will be trained. After applying this operation $k$ times, $H^k$ becomes a node properties vector that contains k-hop local structure information.
		
		Note that, during the repetition of Equation \ref{eq:eq1}, with the exception of $H^k$, the result in every step can only be used to generate the next convolution result. During this process, a large amount of information will be lost, and only the last convolution result $H^k$, which represents the largest sub-graph, could be used for later tasks. This kind of operation can cause a significant loss of information. Only the $k$-hop local structure would be captured by the convolutional layer. Our attention convolution layer aims to solve this issue by attentionally aggregating the information from each convolution step. The comparison of two graph convolution layers is shown in Figure \ref{fig:traditional}.

	    \subsection{Our Proposed Attention Graph Convolution (AGC)} The vast majority of graph neural networks are currently driven by Equation \ref{eq:eq1} which employs k-hop message aggregation mechanism. This enables the node representation to capture the local structural information of k-hop neighbors, but as the number of layers increases, a large amount of early information is lost during each convolution step, which severely affects the final prediction output and also limits the capacity of the model. The core idea of our attention graph convolution (AGC) layer is to enhance the model to not only depend on the k-hop convolution result, but also to capture valuable information from every single hop. The convolution result will thus be a hierarchical representation containing the most valuable information from different hop convolution processes. We exhibit attention behavior and implement it on Equation \ref{eq:eq1} to form a hierarchical node representation $\gamma_{v_n}$ as below:  
		
		\begin{eqnarray}
		    \gamma_{v_n} = \sum_{i=1}^{k} \alpha_iH^k_{v_n}
			\label{eq:eq2}
		\end{eqnarray}
		
    	\begin{figure}
 	 	\centering
 	 		\includegraphics[width=0.5\textwidth]{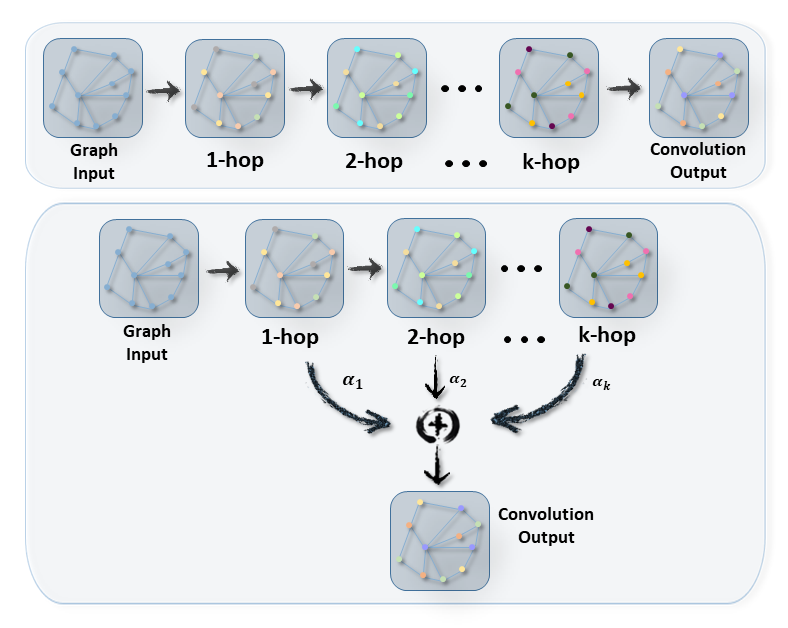}
 	 		\caption{Traditional Graph Convolution Layer (up): Only the final output which contains the largest sub-structure (k-hop neighbor substructure) is used. Attention Graph Convolution Layer (down): valuable information is extracted from every convolution step to generate a hierarchical node representation.}   
 	 	\label{fig:traditional}
		\end{figure}
		
		For simplicity, we use vanilla attention to identify the importance of each hop's aggregation result, in which $\alpha$ is the attention weight and $H^k_{v_n}$ represents node $v_n$'s local structure in $k-hops$. The final node representation contains the hierarchical structure information. Figure \ref{fig:traditional} compares the traditional convolution layer and the attention convolution layer.
		
		To maximize the advantages of deep learning and learn deeper latent features, we use the Residual Learning technique \cite{he2016deep} to stack $m$ attention convolution layers and develop an attention graph convolutional module to obtain a better final node representation $\gamma_{v_n}$.  The input of each AGC layer is the sum of the previous layer's output and the original $X$. Lastly, we use a dense layer to process the combination of outputs from each convolution layer, illustrated as the Attention Graph Convolution Module in Fig \ref{fig:architecture}.
				
		\begin{eqnarray}
		    \gamma_{v_n}^{m+1} = \sum_{i=1}^{k} \alpha_iH^k_{v_n} \qquad H^0_{v_n} = \gamma_{v_n}^{m} + X
		    \label{eq:eq3a}
	    \end{eqnarray}
		\begin{eqnarray}
			\gamma_{v_n} = \text{Dense}(\lbrace\gamma_{v_n}^0 ,\gamma_{v_n}^1,  ... , \gamma_{v_n}^m\rbrace, \theta)
			\label{eq:eq3}
		\end{eqnarray}
		
		where $Dense()$ is a dense layer that combines the outputs from every attention graph convolution layer.  We now have the node representation $\gamma$ for all vertices $v\in G$. For simplicity, we denote the graph as a matrix $G$ with size $n$-by-$c$ where each row is a node's representation.
		
		$$G = (\gamma_{v_1}, \gamma_{v_2}, ..., \gamma_{v_n})$$	
		
		\subsection{Self Attention Pooling}
\begin{figure}[b]
 	 	\centering
 	 		\includegraphics[width=0.5\textwidth]{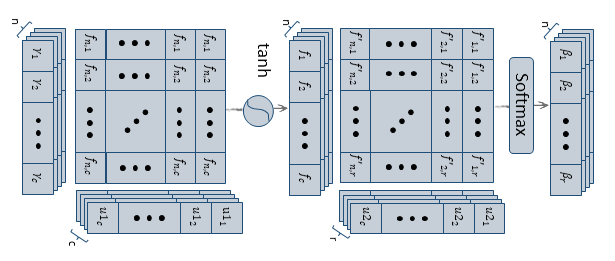}
 	 		\caption{Process of generating Self-Attention Pooling coefficient matrix.}   
 	 	\label{fig:pooling}
		\end{figure}

To perform graph classification task, we would like to generate the graph-level representation from the node's representation. Most previous works use mean/max pooling \cite{henaff2015deep} or sort pooling \cite{defferrard2016convolutional,zhang2018end} to generate a graph representation vector by aggregating all node representation vectors. We believe that simple max/mean pooling or pooling after the sort is ineffective and unnecessary, and therefore propose a self-attention pooling layer as a replacement. The goal is to encode an arbitrary graph into a fixed size embedding matrix while maximizing the information underlying the nodes' representation. Figure \ref{fig:pooling} presents a sample model showing how a coefficient matrix is generated for the attention pooling layer.
		
		We use the attention mechanism by taking the graph node representation learned from the convolution module as the input to output the weights vector $\alpha$. 
		
		\begin{equation}
			\beta = \text{softmax}(u_2 \text{tanh}(u_1 G^T))
			\label{eq:alpha}			
		\end{equation}
		
		In this equation, $u_1$ and $u_2$ are weight matrices with the shape of $c$-by-$c$ and $c$-by-$r$ respectively, where $r$ is a hyperparameter that we set for the number of subspaces to learn the graph representation from the node representation. When $r \geq 1$, $\alpha$ becomes a weight matrix instead of a vector, and Equation \ref{eq:alpha} can then be written as
		
		\begin{equation}
			B = \text{softmax}(u_2 \text{tanh}(u_1 G^T))
			\label{eq:alpha2}			
		\end{equation}
		
		Each row of $B$ represents one node's weight in a different sub-space. The $softmax$ function is performed along the second dimension of its input. We then conduct a weighted summation according to $B$ from Equation \ref{eq:alpha2} to obtain the graph representation matrix $M$ with shape $n$-by-$r$.
		
		\begin{equation}
		 M = B \dot G
		 \label{eq:eq4}
		\end{equation}
		
		\begin{algorithm}[!h]
            \caption{Procedure of DAGCN} 
            \label{alg1}
            \begin{algorithmic}[1]
                \REQUIRE ~~\\
                    $T$: Iterations for updating. \\
                    $A$: Unweighted adjacency matrix; \\
                    $v_n$: Feature vector of node $V_n$ \\
                    $K$: The number of hops for convolution operation; \\
                    $M$: The number of attention graph convolution layers 
                \ENSURE  ~~\\
                    $Y$: Prediction outcome. 
                \STATE Model initialization. $k, m \Leftarrow 0$,  $H_{v_n}^0 \Leftarrow v_n$
                \FOR{ iterator = 1, 2, 3, ..., $T$}
                    \FOR{$m = 1$ \textbf{to} $M$}
                        \FOR{$k = 1$ \textbf{to} $K$}
                            \STATE K-hop graph convolution. Equation (\ref{eq:eq1})
                        \ENDFOR
                        \STATE Attention aggregation from each hop. Equation (\ref{eq:eq2})
                        \STATE Store the result and prepare input for next layer. \\
                           $\gamma_{v_n}^m \Leftarrow H_{v_n}, H_{v_n}^0 \Leftarrow H_{v_n} + X$
                    \ENDFOR
                    \STATE Generalize final node representation $\gamma_{v_n}$ for node $v_n         \in g$. Equation (\ref{eq:eq3}). 
                    \STATE Generalizing coefficient matrix for attention pooling layer.  Equation (\ref{eq:alpha2})
                    \STATE Weighted sum over graph $g$. Equation (\ref{eq:eq4})
                    \STATE Update the all weight parameters with stochastic gradient.
                \ENDFOR
                \STATE \textbf{return} $Y \in \mathbb{R}^{n \times |C|}$  Equation (\ref{eq:eq5})
            \end{algorithmic}
        \end{algorithm}
		
		We now have a graph representation matrix in which each row is a graph representation in one sub-space, and the overall matrix produces a comprehensive representation for the graph. Lastly, a fully-connected layer followed by a softmax layer takes $M$ as the input to accomplish the graph classification. 
		
		\begin{eqnarray}
			Y = softmax(ZM + C)
			\label{eq:eq5}
		\end{eqnarray}
		
		We thus obtain the final classification result $Y$. The step algorithm is summarized in Algorithm \ref{alg1}
		
	\section{Experiments and Results} We construct two sets of experiments to evaluate DAGCN with both graph kernel and GCNs methods in a graph classification task. Both experiments are based on several popular benchmark datasets. The reported result shows that DAGCN outperforms the state-of-the-art deep learning methods and yields a competitive result compared to graph kernels. Details of the code and data are available at https://github.com/dawenzi123/DAGCN

    \subsection{Datasets \& Baselines}We use seven benchmark bioinformatics datasets to evaluate our DAGCN model according to the accuracy of the graph classification task. The datasets used are: NCI1, D\&D, ENZYMES, NCI109, PROTEINS and PTC. Brief data information is listed in Table \ref{tb:dataset}, and a detailed dataset description can be found in \cite{yanardag2015deep}. For the baselines, we compare our framework with major families of graph kernels in the literature and some newly deep learning approaches. For the Graph Kernel Baselines, we compare DAGCN with five state-of-the-art graph kernels: a) Random Walk (RW)\cite{gartner2003graph}, b) Shortest Path Kernel (SP)\cite{borgwardt2005shortest}, c) Graphlet Kernel (GK)\cite{kondor2009graphlet}, d) Weisfeiler-Lehman (WL)\cite{shervashidze2011weisfeiler}, and e) Deep Graph Kernels (DGK)\cite{yanardag2015deep}. In the same benchmark datasets, we also compare our DAGCN model with four deep learning approaches for graph classification. Because of the large amount of literature related to GCN, we could not compare every method. DCNN, PSCN, ECC and DGCNN are four recently proposed state-of-the-art GCNs which are most related to our approach. 

    \begin{table}[ht]
    \caption{}
    \centering
    \scalebox{0.75}{
    \begin{tabular}{cccccccc}
    \hline
        Dataset         & NCI1  & D\&D   & ENZYMES   & MUTAG  & NCI109    & PROTEINS  & PTC     \\ \hline
        Nodes (max)      & 111   & 5748   & 126       & 28     & 111       & 620       & 109     \\
        Nodes (avg.)     & 29.80 & 284.32 & 32.60     & 17.93  & 29.60     & 39.06     & 25.56   \\
        Graphs          & 4110  & 1178   & 600       & 188    & 4127      & 1113      & 344     \\ \hline
    \end{tabular}}
    \label{tb:dataset}
    \end{table}
    
    \subsection{Graph Kernel Configuration} For the graph kernel parameter setting, the height parameters of WL and PK are chosen from the set $\lbrace0, 1, 2, 3, 4, 5 \rbrace$. For the Random Walk (RW) kernel, we set the decay parameter as $\lambda$, following the suggestion in \cite{shervashidze2011weisfeiler}. Results for the others were borrowed from previous works \cite{niepert2016learning,yanardag2015deep,zhang2018end}. All the experimental setups were the same so that a fair comparison could be made.
    
    For PSCN, ECC and DGCNN, we adopted the best results from the paper \cite{verma2018graph}, since their experiment settings are the same as ours. For DCNN, we conducted the experiment based on the standard setting discussed below. For fairness, we also removed the edge features from all datasets, as most of the graph data were missing edge features and the methods we compared do not leverage edge features.  
    
    We attempted not to fine tune our model to improve performance. The same configuration with rough default values were shared between two sets of experiences. The hidden layer size for all dense layers and convolution layers was set to 64, $k$ was chosen from sets $\lbrace 1, 5, 10 \rbrace$, and the chosen number of hops was $k \in \lbrace 3, 5, 10 \rbrace$. For the general setting, we adopted the same procedure as previous works \cite{zhang2018end} so that a fair comparison could be made. We used the Adam \cite{kingma2014adam} optimization policy with L2 regularization and learning rate selected from $\lbrace 0.01, 0.001, 0.0001 \rbrace$ to ensure the best play of the model. The batch size was fixed as 50, and 10-fold cross validation was implemented (9 folds for training, 1 fold for testing) to report the average classification accuracy and standard deviations.
	
	\begin{table*}[t]
        \centering
        \caption{Comparison with deep learning methods}
        \begin{tabular}{cccccccc}
\hline
Dataset              & NCI1                             & ENZYMES                 & MUTAG                   & NCI109                  & PROTEINS                & PTC     \\
\hline
DCNN                 & 56.61$\pm$1.04                  & 42.44$\pm$1.76          & -                       & 57.47$\pm$1.22          & 61.29$\pm$1.60          & 56.60$\pm$2.89 \\
PSCN                 & 76.34$\pm$1.68                       & -                       & -                       & -                       & 75.00$\pm$2.51          & 62.29$\pm$5.68 \\
ECC                  & 76.82                                   & 45.67                   & -                       & 75.03                   & -                       & -              \\
DGCNN                & 74.44$\pm$0.47            & 51.00$\pm$7.29          & 85.83$\pm$1.66          & 75.03$\pm$1.72          & 75.54$\pm$0.94          & 58.59$\pm$2.47 \\
DAGCN               & \textbf{81.68$\pm$1.69}       & \textbf{58.17$\pm$8.76} & \textbf{87.22$\pm$6.1}  & \textbf{81.46$\pm$1.51} & \textbf{76.33$\pm$4.3}  & \textbf{62.88$\pm$9.61} \\ 
\hline
\multicolumn{1}{l}{} & \multicolumn{1}{l}{} & \multicolumn{1}{l}{} & \multicolumn{1}{l}{} & \multicolumn{1}{l}{} & \multicolumn{1}{l}{} & \multicolumn{1}{l}{} &               
        \end{tabular}
    \label{tb:result1}
    \end{table*}

\begin{table*}[t]
\centering
\caption{Comparison with graph kernels}
\begin{tabular}{cccccccc}
\hline
Dataset              & NCI1   & ENZYMES                 & MUTAG                   & NCI109                  & PROTEINS                & PTC     \\
\hline
RW           & -                    & 24.16$\pm$1.64       & 79.17$\pm$2.07       & \textgreater 1 Day   & 74.22$\pm$0.42       & 57.85$\pm$1.30 \\
SP                   & 73.00$\pm$0.24       & 40.10$\pm$1.50       & -                    & 73.00$\pm$0.24       & 75.07$\pm$0.54       & 58.24$\pm$2.44 \\
GK                   & 62.28$\pm$0.29          & 26.61$\pm$0.99       & 81.39$\pm$1.74       & 62.60$\pm$0.19       & 71.67$\pm$0.55       & 57.26$\pm$1.41 \\
WL                   & \textbf{82.19$\pm$0.18}             & 52.22$\pm$1.26       & 84.11$\pm$1.91       & \textbf{82.46$\pm$0.24}       & 74.68$\pm$0.49       & 57.97$\pm$0.49 \\
DGK                  & 80.31$\pm$0.46          & 53.43$\pm$0.91       & -                    & 80.32$\pm$0.33       & 75.68$\pm$0.54       & 60.08$\pm$2.55 \\
DAGCN               & 81.68$\pm$1.69          & \textbf{58.17$\pm$8.76}       & \textbf{87.22$\pm$6.1}        & 81.46$\pm$1.51       & \textbf{76.33$\pm$4.3}        & \textbf{62.88$\pm$9.61} \\ 
\hline
\multicolumn{1}{l}{} & \multicolumn{1}{l}{} & \multicolumn{1}{l}{} & \multicolumn{1}{l}{} & \multicolumn{1}{l}{} & \multicolumn{1}{l}{} & \multicolumn{1}{l}{} &               
\end{tabular}
\label{tb:result2}
\end{table*}

	\subsection{Experimental Result} Table \ref{tb:result1} shows the average classification accuracy of the compared deep learning methods.``$-$'' in the table means that either the source code is not available or the previous report did not contain a related result. From the results, we can see that our proposed model consistently outperforms all other methods on six of the seven datasets, and is second best on D\&D. In particular, there is a 7\% improvement in classification accuracy on NCI1 and more than 8\% on NCI109, with a 1\% - 3\% accuracy gain on the other four datasets (excluding D\&D). DAGCN outperforms DCNN and ECC in every case, proving our hypothesis that simple summing the node features is ineffective and will result in the loss of topology information. PSCN performs about the same as our model on PROTEINS and PTC but is much worse on NCI1 because it is more likely to overfit predefined node ordering. We avoid this problem by using attention pooling which dynamically learns the valuable node distributions over the graph. The improvement achieved by DAGCN can be explained as follows. 1) By using an attention mechanism to aggregate different hop neighbors, DAGCN is able to access more information underlying the graph input, thus achieving better performance. 2) By using the attention pooling layer, DAGCN is able to capture multiple graph signatures on the fly without losing any individual node or global topology information. 
	
	We also compare DAGCN with state-of-the-art graph kernels. The result in table \ref{tb:result2} show that DAGCN is very competitive with state-of-the-art graph kernels. Our model is consistent among the top-2 in terms of performance on all datasets. This is a 1\% - 3\% improvement in accuracy on most datasets, with a high of 9\% improvement for ENZYMES, compared with graph kernels other than WL.
    
	\section{Case Study}
\begin{figure*}
 	 	\centering
 	 		\includegraphics[width=0.8\textwidth]{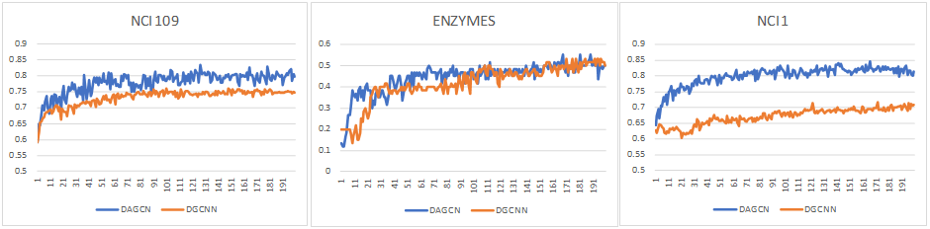}
 	 	    \caption{Learning curve for DAGCN (blue) and DGCNN (orange)}   
 	 	\label{fig:efficiency}
	\end{figure*}

	The experiment results clearly demonstrate the classification performance of DAGCN compared with other deep learning GCNs. We also compare the efficiency of DAGCN with one of the most recent deep learning models, DGCNN, on NCI1, ENZYMES and NCI109, three benchmark datasets on which the learning process is observed to be relatively stable. Since the most significant learning process occurs in the early stages of training, we set the iteration number for both models on all datasets to 200. Although DAGCN has a Residual Learning structure to enhance performance, we limit the number of attention graph convolution layers $m$ to 1 to make this comparison fair. The learning rate has the same setting as DGCNN's default, and all other parameters have the default setting previously mentioned. Figure \ref{fig:efficiency} shows that DAGCN not only achieves better classification accuracy, but also has a better rate of convergence.
	
	Compared with deep learning methods, DAGCN has obvious advantages over graph kernels. Although the overall state-of-the-art in the graph classification task is still dominated by graph kernels, DAGCN is the most practical in its ability to address efficiency and several other issues which most graph kernels suffer from.
	
	\textbf{Computational complexity.} Graph kernels first need to compute the similarity between each two graphs in the training dataset to form a similarity matrix. Given a dataset of size $N$, then $N(N-1)/2$ computation steps are required. This number will grow exponentially when the size of the dataset is increased. In addition, calculating the similarity between a pair of graphs is also an exponential operation based on the number of nodes in the graph. This limits the power of the graph kernels only working for small data-set with small graph. By design, the computational complexity of DAGCN grows linearly for both the dataset size and graph size.
	
	\textbf{Static graph features.} Graph kernels can also broadly be divided into two parts. First, a similarity matrix is constructed by the pre-defined kernel function, and a deep learning model then learns the classification rules. The two steps are independent of each other. The first step can be envisaged as human feature engineering, after which, the features are fixed and are not optimized during the training process. Similar datasets might share some common features as a result of common natural properties (i.e., two bio-informatics datasets). But datasets from different fields must have different properties (e.g., social network and protein network). Although our model is also created from two modules, it still an end-to-end model. All parameters will be optimized during the training process giving DAGCN more advantage on generality.
	
	\textbf{Single structure.} Due to their nature, graph kernels can only focus on a certain scope of graph according to their kernel function. As a result, either global structure or local properties are lost. Our attention pooling layer enables us to learn hierarchical structure information that includes both local and global properties.
		
	\section{Conclusion and Future Work} In this paper, we have proposed a novel Dual Attention Graph Convolutional Network (DAGCN) model with the core idea of maximally exploiting the original information underlying the graph input. We used an attention mechanism to address the weakness of traditional GCN models, in which information is largely lost in every convolution step. Our attention convolution layer design is capable of capturing more hierarchical structure information than other models and provides a much more informative representation of both individual nodes and the whole graph.  The attention pooling layer generates a fixed size, comprehensive graph representation matrix by using a self-attention mechanism to focus on the different aspects of graph. The experimental results show that our model outperforms other deep learning methods and most graph kernels in a range of datasets.
	
	In future work, We intend to implement and validate our model on more complex graphs such as EHRs data and social networks. We will also analyze graph convolution in greater depth to discover how information is distributed at different convolution level. Lastly, we observe that it would be better to test a larger number of attention architectures to mimic the nature of the dataset, since our model only employs one basic attention architecture for all datasets.
	
	\section*{Acknowledgment}	
 	This research was funded by the Australian Government through the Australian Research Council (ARC) under grants 1) LP160100630 partnership with Australia Government Department of Health and 2) LP150100671 partnership with Australia Research Alliance for Children and Youth (ARACY) and Global Business College Australia (GBCA). We  acknowledge the support of NVIDIA Corporation and MakeMagic Australia with the donation of GPU used for this research.

\balance

	\bibliographystyle{IEEEtran}
	\bibliography{someGraph}
	
\end{document}